\documentclass[letterpaper]{article} % DO NOT CHANGE THIS
\usepackage{aaai2026}  % DO NOT CHANGE THIS
\usepackage{times}  % DO NOT CHANGE THIS
\usepackage{helvet}  % DO NOT CHANGE THIS
\usepackage{courier}  % DO NOT CHANGE THIS
\usepackage[hyphens]{url}  % DO NOT CHANGE THIS
\usepackage{graphicx} % DO NOT CHANGE THIS
\usepackage{natbib}  % DO NOT CHANGE THIS AND DO NOT ADD ANY OPTIONS TO IT
\usepackage{caption} % DO NOT CHANGE THIS AND DO NOT ADD ANY OPTIONS TO IT
\usepackage{algorithm}
\usepackage{algorithmic}

\usepackage{amsmath}
\usepackage{amssymb}
\usepackage{bm}
\usepackage{amsthm}

\usepackage{multirow}
\usepackage{booktabs} % For formal tables
\usepackage{color}
\usepackage[table]{xcolor}
\usepackage{graphicx}
\usepackage{subfig}
\usepackage{makecell}
\usepackage{xspace}

\newtheorem{lemma}{Lemma}
\newtheorem{theorem}{Theorem}
\newtheorem{myDef}{Definition}

\newcommand{\eg}{\emph{e.g.,}\xspace}

\usepackage{newfloat}
\usepackage{listings}
\DeclareCaptionStyle{ruled}{labelfont=normalfont,labelsep=colon,strut=off} % DO NOT CHANGE THIS
\floatstyle{ruled}
\newfloat{listing}{tb}{lst}{}
\floatname{listing}{Listing}
\title{Hierarchical Frequency-Decomposition Graph Neural Networks for Road Network Representation Learning}
\author{
    Jingtian Ma\textsuperscript{\rm 1, \rm 3},
    Jingyuan Wang\textsuperscript{\rm 1, \rm 2, \rm 3, \rm 4}\thanks{Corresponding author.},
    Leong Hou U\textsuperscript{\rm 5}\\
}
\affiliations{
    \textsuperscript{\rm 1}School of Computer Science and Engineering, Beihang University, Beijing, China\\
    \textsuperscript{\rm 2}School of Economics and Management, Beihang University, Beijing, China\\
    \textsuperscript{\rm 3}MIIT Key Laboratory of Data Intelligence and Management, Beihang University, Beijing, China\\
    \textsuperscript{\rm 4}MOE Engineering Research Center of Advanced Computer Application Technology, Beihang University, China\\
    \textsuperscript{\rm 5}University of Macau, Macau SAR, China
}

\begin{document}

\maketitle

\begin{abstract}

Road networks are critical infrastructures underpinning intelligent transportation systems and their related applications. Effective representation learning of road networks remains challenging due to the complex interplay between spatial structures and frequency characteristics in traffic patterns. Existing graph neural networks for modeling road networks predominantly fall into two paradigms: spatial-based methods that capture local topology but tend to over-smooth representations, and spectral-based methods that analyze global frequency components but often overlook localized variations. This spatial-spectral misalignment limits their modeling capacity for road networks exhibiting both coarse global trends and fine-grained local fluctuations. To bridge this gap, we propose HiFiNet, a novel hierarchical frequency-decomposition graph neural network that unifies spatial and spectral modeling. HiFiNet constructs a multi-level hierarchy of virtual nodes to enable localized frequency analysis, and employs a decomposition–updating–reconstruction framework with a topology-aware graph transformer to separately model and fuse low- and high-frequency signals. Theoretically justified and empirically validated on multiple real-world datasets across four downstream tasks, HiFiNet demonstrates superior performance and generalization ability in capturing effective road network representations.

\end{abstract}

\section{Introduction}

The road network serves as the fundamental component of intelligent transportation systems (ITS), which supports a wide range of traffic-related applications such as traffic forecasting~\cite{li2017diffusion}, trajectory inference~\cite{wang2018inferring}, and urban planning~\cite{wang2018cd}. 
To facilitate these applications, it is crucial to build effective and generalizable representations of road networks.
A widely adopted approach is to model the road network as a graph, where nodes denote road segments and edges reflect topological connectivity. 
Learning expressive representations for such road network graphs remains a core challenge.

Early studies for modeling road networks often rely on random walk-based approaches~\cite{perozzi2014deepwalk, grover2016node2vec}, which generate node sequences through random walks and treat them as sentences to learn node embeddings. While simple and scalable, these methods neglect node attributes and fail to capture structural semantics such as frequency-aware patterns.

To address these limitations, graph neural networks (GNNs) have emerged as powerful tools for graph representation learning. Broadly, existing GNNs can be categorized into two paradigms: \textit{spectral-based} and \textit{spatial-based} approaches. Spectral methods~\cite{bruna2013spectral, defferrard2016convolutional} are grounded in graph signal processing theory, defining graph convolutions via the eigenbasis of the Laplacian to analyze node features in the frequency domain. 
In contrast, spatial-based GNNs~\cite{kipf2016semi, velickovic2017graph, pei2018geom} perform message passing by directly aggregating features from local neighborhoods, enabling efficient and inductive learning on graphs with varying topology. 
More recently, transformer-based architectures have been extended to graph domains~\cite{dwivedi2020generalization, ying2021transformers}, incorporating global attention mechanisms along with structural or positional encodings to capture long-range dependencies beyond local neighborhoods.

\begin{figure}
    \centering
    \includegraphics[width=1\linewidth]{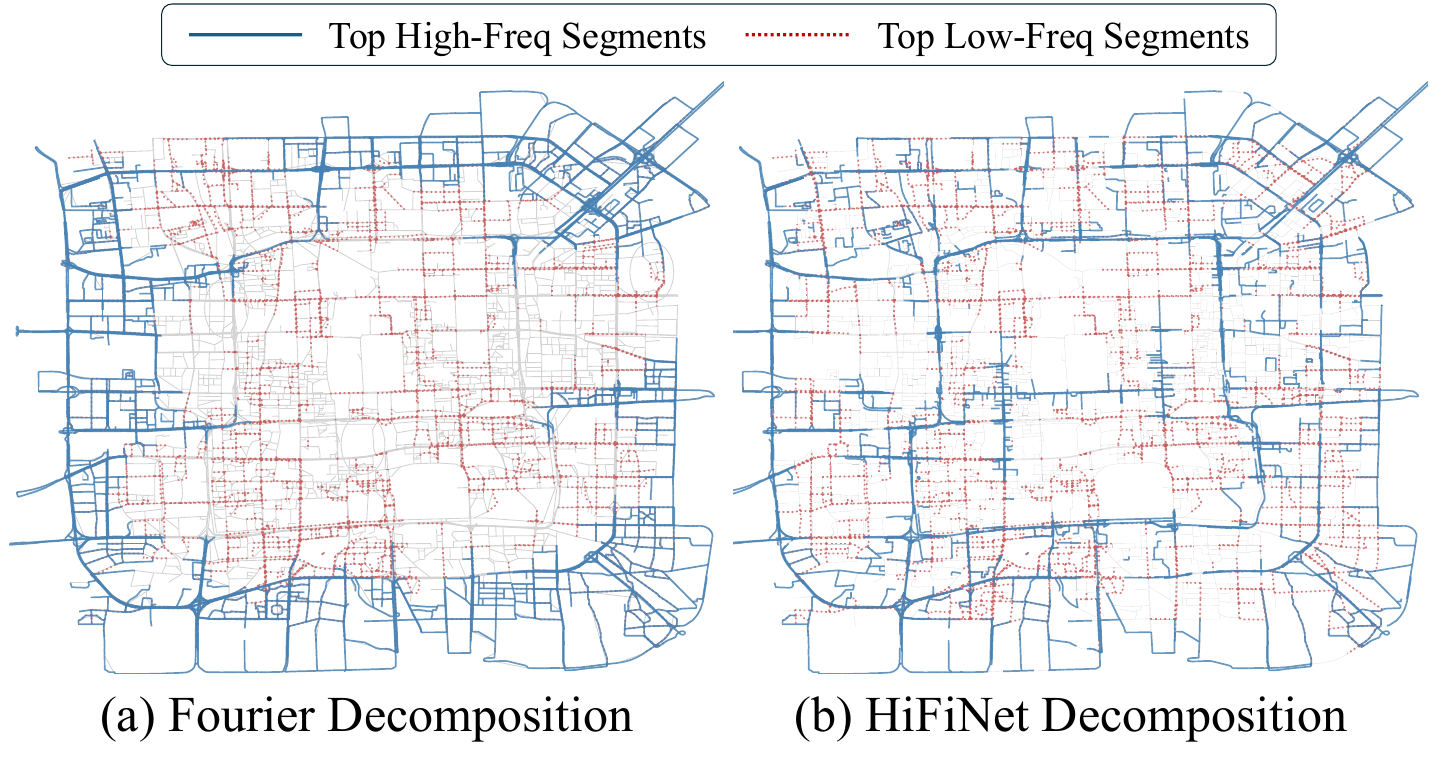}
    \caption{The frequency decomposition of Fourier Transform and HiFiNet on traffic flow signals.}
    \label{fig:intro}
\end{figure}

However, despite these advances, these two paradigms remain largely \textit{spatial-spectral misalignment}. 
Spatial-based models excel at capturing local structural patterns but often act as low-pass filters~\cite{nt2019revisiting,wu2019simplifying,bastos2022expressive}. leading to over-smoothing and poor global expressiveness.
While spectral-based models provide theoretical insights into signal propagation through global frequency decomposition, but they often overlook localized variations.
Consequently, there is a critical gap: a unified framework that simultaneously models spatial structures and frequency components is still lacking, which limits the expressive capacity of current models for road networks that exhibit both coarse global patterns and fine-grained local variations.
As illustrated in Fig.~\ref{fig:intro}(a), we apply the Fourier Transform to traffic flow signals to reveal their frequency characteristics.
Low-frequency edges (blue solid lines) tend to appear in peripheral areas, while high-frequency edges (red dotted lines) are often concentrated in city centers. 
This reflects a typical urban dynamic—peripheral regions are shaped by regular commuting patterns, while central areas exhibit more fluctuation due to diverse land use.
However, exceptions may exist: some peripheral roads maintain stable flow, while certain inner-city segments show significant volatility.
These observations support the presence of spatial-spectral misalignment in existing approaches, where frequency patterns are not well localized, leading to suboptimal representation of complex spatial structures.

To address these challenges, we propose a novel \emph{\underline{Hi}erarchical \underline{F}requency-Decompos\underline{i}tion \underline{Net}work} (\emph{HiFiNet}) for road network representation learning. Our framework integrates spatial and spectral modeling in a unified architecture through two key innovations:
First, we introduce a three-level hierarchy by clustering road segments into \emph{localities} and \emph{regions}, each represented by virtual nodes. This hierarchy not only captures multi-scale spatial semantics but also enables localized graph signal decomposition. 
As shown in Fig.~\ref{fig:intro}(b), high-frequency edges are observed not only at city margins but also along inner ring roads, where traffic flow exhibits significant variability. 
We theoretically prove that this hierarchical construction facilitates the separation of low- and high-frequency components, alleviating over-smoothing and enhancing representational diversity.
Second, we develop a frequency-decomposition learning module following a \emph{decomposition–updating–reconstruction} paradigm. 
It explicitly models low- and high-frequency signals, and updates them via a topology-aware graph transformer, capturing both smooth global trends and sharp local variations. 
These enriched components are then fused into discriminative representations under a unified loss framework that promotes consistency across frequencies and scales.

Our contributions are summarized as follows:
\begin{itemize}
\item We propose \emph{HiFiNet}, a unified spatial-spectral framework that integrates hierarchical graph modeling with localized frequency decomposition for road network representation learning.
\item We design a multi-level hierarchy that not only captures spatial locality but also enables frequency separation, a property we theoretically validate in our framework.
\item We develop a frequency decomposition module that jointly models low- and high-frequency graph signals, enhancing the expressiveness of learned representations.
\item Extensive experiments on real-world datasets across four downstream tasks demonstrate the superior performance and generalization ability of our approach.
\end{itemize}

\section{Related Work}

Our work is related to the following research directions:

\subsubsection{Road Network Modeling.}
Road network modeling focuses on capturing the structural and semantic characteristics of urban road systems. Early methods such as DeepWalk~\cite{perozzi2014deepwalk} and Node2vec~\cite{grover2016node2vec} used random walks to learn shallow representations, but lacked the capacity to incorporate node attributes. With the advent of GNNs, models like RFN~\cite{jepsen2020relational} and Geom-GCN~\cite{pei2018geom} leveraged neighbor aggregation for representation learning. To model long-range dependencies, hierarchical approaches such as HRNR~\cite{wu2020learning} introduced pooling operations. However, most GNN-based models suffer from over-smoothing and often overlook high-frequency components that are critical for preserving fine-grained road patterns.

% \subsubsection{Graph Spectral Theory.}
% Graph spectral methods analyze structural properties through spectral decomposition of matrices like the Laplacian. Classical models such as ChebNet~\cite{defferrard2016convolutional} and GCN~\cite{kipf2016semi} inherently perform low-pass filtering, which aids denoising but causes over-smoothing. Recent studies~\cite{wu2019simplifying, zhu2021interpreting} propose frequency-aware GNNs that balance low- and high-frequency components for better expressiveness. However, most spectral methods focus on node-level tasks, and their potential for modeling hierarchical and functional patterns in road networks remains underexplored.

\subsubsection{Graph Spectral Theory.}
Graph spectral methods analyze structural properties through spectral decomposition of matrices like the Laplacian. Classical models such as ChebNet~\cite{defferrard2016convolutional}, GCN~\cite{kipf2016semi}, and CayleyNet~\cite{levie2018cayleynets} inherently perform low-pass filtering, which aids denoising but causes over-smoothing. Recent studies~\cite{wu2019simplifying, zhu2021interpreting, bo2021beyond} propose frequency-aware GNNs that balance low- and high-frequency components for better expressiveness. However, most spectral methods focus on node-level tasks, and their potential for modeling hierarchical and functional patterns in road networks remains underexplored.

\subsubsection{Graph Neural Networks.}
GNNs have achieved remarkable success in graph representation learning. Representative models include GCN~\cite{kipf2016semi}, GAT~\cite{velickovic2017graph}, and GraphSAGE~\cite{hamilton2017inductive}, which follow a message-passing paradigm. However, these methods tend to behave as low-pass filters, causing over-smoothing~\cite{nt2019revisiting}. To better capture global context, recent work explores Graph Transformers~\cite{dwivedi2020generalization, ying2021transformers, wu2022nodeformer}, which use attention mechanisms to learn long-range dependencies. Yet, how to balance local structural priors and global flexibility remains an open challenge.

\section{Preliminaries}

In this section, we introduce the notations used throughout the paper and formally define our task. 

\begin{myDef}[Road Network]
A road network is modeled as a directed graph $\mathcal{G}=\langle\mathcal{S}, \bm{A}_S\rangle$, where $\mathcal{S}$ denotes the set of $N_S$ road segments, and $\bm{A}_S \in \mathbb{R}^{N_S \times N_S}$ is the binary adjacency matrix, with $\bm{A}_S[i,j] = 1$ indicating a directed connection from segment $s_i$ to $s_j$, and 0 otherwise.
\end{myDef}

% Here, we follow the widely adopted setting \cite{wang2019empowering, li2018multi,jepsen2019graph} by treating road segments as vertices. It is equally feasible to define the locations (\eg~intersection) as vertices. 

\begin{myDef}[Segment Signal]\label{def:sig}
The segment signal matrix $\bm{X}_S \in \mathbb{R}^{N_S \times d_0}$ encodes raw attributes (e.g., road class, lane number, traffic flow), where each row $\bm{x}_S^i$ corresponds to segment $s_i$. From the graph signal processing view, $\bm{X}_S$ can be decomposed into low-frequency components (smooth global patterns) and high-frequency components (local variations).
\end{myDef}

To facilitate frequency-aware hierarchical modeling, we introduce two virtual node types: \textit{localities} and \textit{regions}, and organize the road network as a three-level hierarchy: \emph{segment} $\rightarrow$ \emph{locality} $\rightarrow$ \emph{region}.

\begin{myDef}[Locality]
    A locality $l\in\mathcal{L}$ refers to a group of spatially adjacent road segments that collectively serve a specific traffic-related function (e.g., overpass, intersection), where $\mathcal{L}$ denotes the set of $N_L$ localities.
\end{myDef}

\begin{myDef}[Region]
    A region $r \in \mathcal{R}$ consists of multiple localities and represents a broader urban area with a specific functional role (e.g., residential, commercial zone), where $\mathcal{R}$ denotes the set of $N_R$ regions.
\end{myDef}

% We further formalize the hierarchical road network based on these concepts.

% \begin{myDef}[Hierarchical Road Network] 
%     A hierarchical road network is defined as $\mathcal{H}=\langle \mathcal{V},\mathcal{E}\rangle$, where $\mathcal{V}=\mathcal{S}\cup\mathcal{L}\cup\mathcal{R}$ is the set of nodes including segments, localities, and regions.
%     The edge set is defined as $\mathcal{E}=\{\bm{A}_S, \bm{A}_L, \bm{A}_R, \bm{A}_{SL}, \bm{A}_{LR}\}$, 
%     where the matrices $\bm{A}_S\in\mathbb{R}^{N_S\times N_S}$, $\bm{A}_L\in\mathbb{R}^{N_L \times N_L}$, $\bm{A}_R\in\mathbb{R}^{N_R\times N_R}$, $\bm{A}_{SL}\in\mathbb{R}^{N_S \times N_L}$, and $\bm{A}_{LR}\in\mathbb{R}^{N_L\times N_R}$ denote (binary or weighted) adjacency matrices for capturing the links between (1) two segment nodes, (2) two locality nodes, (3) two region nodes, (4) a segment node and a locality node, and (5) a locality node and a region node, respectively. 
% \end{myDef}
\begin{myDef}[Hierarchical Road Network] 
We define a hierarchical road network as $\mathcal{H}=\langle \mathcal{V}, \mathcal{E} \rangle$, where $\mathcal{V} = \mathcal{S} \cup \mathcal{L} \cup \mathcal{R}$ denote the set of all nodes, and $\mathcal{E} = \{\bm{A}_S, \bm{A}_L, \bm{A}_R, \bm{A}_{SL}, \bm{A}_{LR}\}$ includes adjacency matrices for:
(1) segment–segment ($\bm{A}_S$), 
(2) locality–locality ($\bm{A}_L$), 
(3) region–region ($\bm{A}_R$),
(4) segment–locality ($\bm{A}_{SL}$), and 
(5) locality–region ($\bm{A}_{LR}$) relations.
\end{myDef}

Unlike segment adjacency matrix, the matrices $\bm{A}_L$, $\bm{A}_R$, $\bm{A}_{SL}$, and $\bm{A}_{LR}$ are treated as learnable associations. In particular, $\bm{A}_{SL}$ and $\bm{A}_{LR}$ act as \emph{segment-to-locality} and \emph{locality-to-region} assignment matrices that support hierarchical aggregation. With these definitions, we are now ready to formally define our task.

\begin{myDef}[Road Network Representation Learning]
Given a road network $\mathcal{G}$ and segment signal matrix $\bm{X}_S$, the objective is to construct the hierarchical road network $\mathcal{H}$ and learn a $d$-dimensional embedding $\bm{h}_m \in \mathbb{R}^d$ for each node $m \in \mathcal{V}$, where $d \ll |\mathcal{V}|$. The learned representations are expected to preserve both low- and high-frequency semantics and generalize across downstream traffic tasks.
\end{myDef}
\section{Hierarchical Frequency-Decomposition Network}

% In this section, we propose the \emph{\underline{Hi}erarchical \underline{F}requency-Decompos\underline{i}tion \underline{Net}work} (\emph{HiFiNet}) for road network representation learning. The overall framework is shown in Fig.~\ref{fig:framework}. We begin by constructing a three-level hierarchical architecture to capture the multi-scale structural properties of road networks. We provide a theoretical analysis demonstrating that this hierarchical design facilitates the effective decomposition of \emph{low-frequency} and \emph{high-frequency} components in road network signals. We then explicitly model these two components and integrate them to learn the final road network representations. By jointly capturing both smooth and variant information, HiFiNet produces comprehensive and semantically meaningful representations of road networks. 

\begin{figure}
    \centering
    \includegraphics[width=1\linewidth]{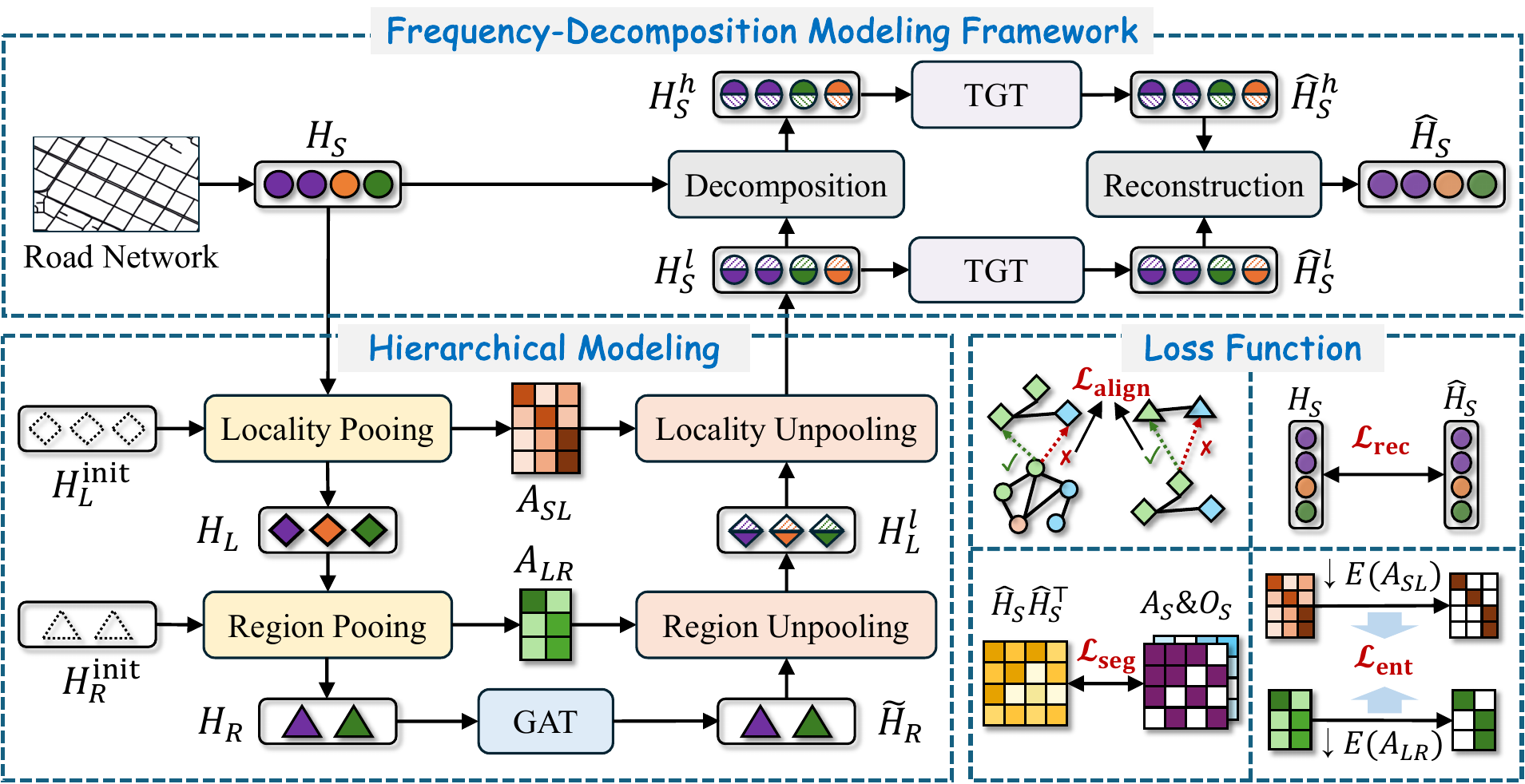}
    \caption{The overall framework of HiFiNet.}
    \label{fig:framework}
\end{figure}

In this section, we propose the \emph{\underline{Hi}erarchical \underline{F}requency-Decompos\underline{i}tion \underline{Net}work} (\emph{HiFiNet}) for road network representation learning. As illustrated in Fig.~\ref{fig:framework}, HiFiNet constructs a three-level hierarchical architecture to capture the multi-scale structure of road networks. We theoretically demonstrate that this design enables effective decomposition of \emph{low-frequency} and \emph{high-frequency} components in road signals. These components are then explicitly modeled and fused to produce the final representations that capture both smooth and variant patterns of the road network.

\subsection{Hierarchical Architecture Modeling}

\subsubsection{Contextual Embedding for Road Segments.}

Road segments are often associated with rich contextual attributes. 
To incorporate this auxiliary information, we map these attributes to latent embedding vectors for individual segments.

Given a segment $s_i$, its raw attribute vector $\bm{x}_S^i \in \mathbb{R}^{d_0}$ (Def.~\ref{def:sig}) includes four key attributes: segment ID, lane number (LN), segment length (SL), and geographical location (longitude and latitude, denoted as LL). 
We map each attribute (or its discretized bin) to a learnable embedding, denoted as $\bm{e}_{ID}^i$, $\bm{e}_{LN}^i$, $\bm{e}_{SL}^i$, and $\bm{e}_{LL}^i$, respectively. The resulting contextual embedding of $s_i$ is:
\begin{equation}
    \bm{v}_S^i = \bm{e}_{ID}^i \Vert \bm{e}_{LN}^i \Vert \bm{e}_{SL}^i \Vert \bm{e}_{LL}^i,
\end{equation}
where ``$\Vert$'' denotes vector concatenation.

Stacking the contextual embeddings of all segments yields the matrix $\bm{V}_S \in \mathbb{R}^{N_S \times d'}$, where $d'$ is the total dimension of the concatenated embeddings. We then apply a two-layer feed-forward network (FFN) with nonlinearity to obtain the initial segment feature matrix:
\begin{equation}\label{eq:HS}
    \bm{H}_S = \mathrm{FFN}(\bm{V}_S),
\end{equation}
where $\bm{H}_S \in \mathbb{R}^{N_S \times d}$ denotes the initial segment features used in the subsequent hierarchical modeling.

\subsubsection{Locality Graph Construction.}
Localities serve as clusters of segments to capture local connectivity patterns such as intersections. A key component is the \emph{segment-to-locality} assignment matrix $\bm{A}_{SL}$, which encodes how segments are grouped into localities based on structural similarity.

We assume that each segment is softly assigned to a locality, with contributions weighted by importance. To model this varying importance, we employ a cross-attention mechanism to capture the interactions between segments and localities. Let $\bm{H}_L^{\mathrm{init}} \in \mathbb{R}^{N_L \times d}$ be a randomly initialized learnable matrix representing locality embeddings. The assignment matrix is computed as
\begin{equation}\label{eq:ASL}
    \bm{A}_{SL}=\mathrm{softmax}\left(\frac{(\bm{H}_S\bm{W}_S)(\bm{H}_L^{\mathrm{init}}\bm{W}_L)^\top}{\sqrt{d}}\right),
\end{equation}
where $\mathrm{softmax}(\cdot)$ denotes the row-wise normalization, and $\bm{W}_S$, $\bm{W}_L$ are learnable projection matrices.
Each entry in $\bm{A}_{SL}$ models the conditional probability of assigning segment $s_i$ to locality $l_j$: $\bm{A}_{SL}[i,j] = \mathrm{Pr}(l_j|s_i)$.

Given $\bm{A}_{SL}$, we compute locality features by aggregating segment features with residual connections:
\begin{equation}\label{eq:HL}
    \bm{H}_L = \bm{A}_{SL}^\top\bm{H}_S + \bm{H}_L^{\mathrm{init}}.
\end{equation}
The locality adjacency matrix is constructed as:
\begin{equation}\label{eq:AL}
    \bm{A}_L = \bm{A}_{SL}^\top\bm{A}_S\bm{A}_{SL},
\end{equation}
which can be interpreted as
\begin{equation} \label{eq-ar-exp}
    \bm{A}_L[i, j]=\sum_{s_m,s_n}\mathrm{Pr}(l_i|s_m)\cdot\mathrm{Pr}(l_j|s_n)\cdot\bm{A}_S[m,n].
\end{equation}

\subsubsection{Region Graph Construction.}
Urban areas often consist of regions with distinct functional roles~\cite{10.1145/2339530.2339561}. We build a region-level graph over localities using a learnable \emph{locality-to-region} assignment matrix $\bm{A}_{LR}$.

Let $\bm{H}_R^{\mathrm{init}} \in \mathbb{R}^{N_R \times d}$ denote the initial region embeddings. The assignment matrix is computed similarly via attention:
\begin{equation}\label{eq:ALR}
    \bm{A}_{LR}=\mathrm{softmax}\left(\frac{(\bm{H}_L\bm{W}_L)(\bm{H}_R^{\mathrm{init}}\bm{W}_R)^\top}{\sqrt{d}}\right),
\end{equation}
where $\bm{W}_L$ and $\bm{W}_R$ are learnable projection matrices and each entry $\bm{A}_{LR}[j,k] = \Pr(r_k|l_j)$.

The region features and adjacency matrix are constructed by aggregating locality features and connections:
\begin{equation}
    \bm{H}_R=\bm{A}_{LR}^\top\bm{H}_L+\bm{H}_R^{\mathrm{init}}, \quad
    \bm{A}_R=\bm{A}_{LR}^\top\bm{A}_L\bm{A}_{LR},
\end{equation}

We theoretically demonstrate that our hierarchical structure exhibits favorable spectral properties. Specifically:

\begin{theorem}
Let \( A_{XY} \in \mathbb{R}^{N_Y \times N_X} \) denote an assignment matrix satisfying the equi-partition and row-normalization properties. Then, the projection of graph signals from the original graph \( X \) to the coarsened graph \( Y \) approximately preserves the low-frequency energy while attenuating high-frequency components.
\end{theorem}

This theorem indicates that the proposed hierarchical projection naturally acts as a spectral low-pass filter, preserving smooth signal components while suppressing high-frequency noise. 
Proofs are provided in Appendix~A.

\subsubsection{Low-frequency Feature Propagation.}
Once the \emph{segment-locality-region} hierarchical structure is established, we adopt a top-down message propagation strategy to propagate these preserved low-frequency features.

We begin by applying a standard Graph Attention Network (GAT)~\cite{velickovic2017graph} to update the node features in the region-level graph, allowing each region node to capture coarse-grained global contextual information:
\begin{equation}
    \tilde{\bm{H}}_R = \mathrm{GAT}(\bm{H}_R, \bm{A}_R).
\end{equation}

Next, since regions aggregate multiple localities, they capture coarse-grained low-frequency patterns. We propagate this information to localities via the assignment matrix $\bm{A}_{LR}$, followed by a GAT module for refinement:
\begin{equation}
    \bm{\tilde{H}}_L^{l} = \bm{A}_{LR}\bm{\tilde{H}}_R, \quad \bm{H}_L^{l} = \mathrm{GAT}(\bm{\tilde{H}}_L^{l}, \bm{A}_L),
\end{equation}
where $\bm{H}_L^{l}\in\mathbb{R}^{N_L\times d}$ is the low-frequency locality features.

Similarly, segment-level low-frequency features are obtained by aggregating from localities via $\bm{A}_{SL}$:
\begin{equation}
    \bm{\tilde{H}}_S^{l} = \bm{A}_{SL}\bm{H}_L^{l}, \quad \bm{H}_S^{l} = \mathrm{GAT}(\bm{\tilde{H}}_S^{l}, \bm{A}_S),
\end{equation}
where $\bm{H}_S^{l}\in\mathbb{R}^{N_S\times d}$ is the low-frequency segment features.

This top-down unpooling process effectively suppresses high-frequency fluctuations and retains low-frequency components, providing essential input for the subsequent frequency decomposition modeling module.

\subsection{Frequency Decomposition Modeling}
We adopt a \emph{decomposition-updating-reconstruction} framework to explicitly separate and model the low-frequency and high-frequency components of segment features, which enables more effective representation learning.

\subsubsection{Decomposition Stage.} 
Since the original segment features $\bm{H}_S$ obtained by Eq.~\eqref{eq:HS} contain both low-frequency and high-frequency components, we can obtain the high-frequency part by subtracting the low-frequency signal:
\begin{equation}
    \bm{H}_S^{h} = \bm{H}_S - \bm{H}_S^{l},
\end{equation}
where $\bm{H}_S, \bm{H}_S^l, \bm{H}_S^h \in \mathbb{R}^{N_S \times d}$ denote the original, low-frequency, and high-frequency segment features separately.

\subsubsection{Updating Stage.}
As road segment graphs often involve a large number of nodes, limiting aggregation to local neighborhoods may fail to capture long-range dependencies, leading to over-smoothing. We thus propose a topology-aware graph transformer (TGT) that integrates global attention with local structure to update both frequency components.

Taking the low-frequency features as an example, we have
\begin{equation}
    \tilde{\bm{H}}_S^l = \mathrm{TGT}(\bm{H}_S^l, \bm{A}_S),
\end{equation}
where $\tilde{\bm{H}}_S^l \in \mathbb{R}^{N_S \times d}$ is the updated low-frequency feature matrix, and $\bm{A}_S$ is the segment-level adjacency matrix.

The TGT module consists of $N$ blocks. We set $\bm{H}_S^{l,0} = \bm{H}_S^l$, and for each block $i = 0, \dots, N-1$, we compute
\begin{equation}
    \bm{Q}_{l,i}=\bm{H}_S^{l,i}\bm{W}_q^{l,i}, \quad \bm{K}_{l,i}=\bm{H}_S^{l,i}\bm{W}_k^{l,i}, \quad \bm{V}_{l,i}=\bm{H}_S^{l,i}\bm{W}_v^{l,i},
\end{equation}
where $\bm{W}_q^{l,i}$, $\bm{W}_k^{l,i}$, and $\bm{W}_v^{l,i}$ are learnable matrices.

Next, we integrate global attention and local topology via
\begin{equation}
    \bm{\mathrm{ATT}}_{l,i} = \alpha\cdot\mathrm{softmax}\left(\frac{\bm{Q}_{l,i}\bm{K}_{l,i}^\top}{\sqrt{d}}\right) + (1-\alpha)\cdot\bm{A}_S,
\end{equation}
where $\alpha$ is a learnable parameter balancing global and local information. The features are updated as
\begin{equation}
    \tilde{\bm{H}}_S^{l, i}=\mathrm{LayerNorm}\left(\bm{\mathrm{ATT}}_{l,i}\bm{V}_{l,i} + \bm{H}_S^{l, i}\right),
\end{equation}
\begin{equation}
    \bm{H}_S^{l, i+1}=\mathrm{LayerNorm}\left(\mathrm{FFN}(\tilde{\bm{H}}_S^{l, i})+\tilde{\bm{H}}_S^{l, i}\right),
\end{equation}
% where $\mathrm{LayerNorm}(\cdot)$ represents the layer normalization operation and $\mathrm{FFN}(\cdot)$ denotes a two-layer feed-forward network.
where $\mathrm{LayerNorm}(\cdot)$ denotes the layer normalization operation.
After processing all blocks, we obtain $\tilde{\bm{H}}_S^l=\bm{H}_S^{l, N}$.

Similarly, the high-frequency features are updated through the same TGT process, yielding 
\begin{equation}
    \tilde{\bm{H}}_S^h = \mathrm{TGT}(\bm{H}_S^h, \bm{A}_S).
\end{equation}

\subsubsection{Reconstruction Stage.}
After updating both components, we reconstruct the final segment features by combining the low-frequency and high-frequency signals:
\begin{equation}
    \hat{{\bm{H}}}_S = \beta\cdot\tilde{\bm{H}}_S^l + (1-\beta)\cdot\tilde{\bm{H}}_S^h,
\end{equation}
where $\hat{\bm{H}}_S \in \mathbb{R}^{N_S \times d}$ is the reconstructed segment feature matrix, and $\beta$ is a learnable parameter that balances the contributions of low-frequency and high-frequency information.

\subsection{Model Training}
Our model involves various learnable parameters, including node representations~(\eg~$\bm{H}_*$), assignment matrices~(\eg~$\bm{A}_*$) within the hierarchical structure, and trainable parameters for frequency decomposition modeling. 
To jointly optimize these components while satisfying theoretical constraints, we design a set of tailored loss functions that jointly optimize the model.

\subsubsection{Alignment Loss.} 
Since the hierarchical structure performs node aggregation, the feature of each child node should closely resemble that of its corresponding parent, while remaining distinct from non-parent nodes. We design a contrastive loss to encourage this property:
\begin{equation}
    \mathcal{L}_{\mathrm{align}}^{SL} = -\frac{1}{N_S} \sum_{i=1}^{N_S} \log \left(
    \frac{
        \exp \left( \mathrm{sim}(\bm{h}_S^i, \bm{h}_L^{p(i)}) / \tau \right)
    }{
        \sum_{j=1}^{N_L} \exp \left( \mathrm{sim}(\bm{h}_S^i, \bm{h}_L^j) / \tau \right)
    } \right),
    \nonumber
\end{equation}
\begin{equation}
    \mathcal{L}_{\mathrm{align}}^{LR} = -\frac{1}{N_L} \sum_{j=1}^{N_L} \log \left(
    \frac{
        \exp \left( \mathrm{sim}(\bm{h}_L^j, \bm{h}_R^{p(j)}) / \tau \right)
    }{
        \sum_{k=1}^{N_R} \exp \left( \mathrm{sim}(\bm{h}_L^j, \bm{h}_R^k) / \tau \right)
    } \right),
    \nonumber
\end{equation}
\begin{equation}
    \mathcal{L}_{\mathrm{align}} = \frac{1}{2} \left( \mathcal{L}_{\mathrm{align}}^{SL} + \mathcal{L}_{\mathrm{align}}^{LR} \right),
\end{equation}
where $\bm{h}_S^i$, $\bm{h}_L^j$, and $\bm{h}_R^k$ denote the features of the $i$-th segment, $j$-th locality, and $k$-th region, respectively; $p(i)$ and $p(j)$ represent the parent node indices; $\tau$ is a temperature parameter controlling distribution sharpness; and $\mathrm{sim}(\cdot, \cdot)$ denotes the cosine similarity function.

\subsubsection{Reconstruction Loss.} 
To ensure the frequency decomposition module effectively retains key information, we require that the reconstructed segment features remain consistent with the original segment features:
% To this end, we design a reconstruction loss function that explicitly enforces this consistency during training.
\begin{equation}
    \mathcal{L}_{\mathrm{rec}} = \frac{1}{N_S} \sum_{i=1}^{N_S} \left\| \hat{\bm{h}}_S^i - \bm{h}_S^i \right\|_2^2,
\end{equation}
where $\hat{\bm{h}}_S^i$ and $\bm{h}_S^i$ denote the reconstructed and original features of the $i$-th segment, respectively.

\subsubsection{Semantic Loss.} To ensure that the reconstructed segment features capture the semantic structure of the road network, we align their pairwise similarities with the relational structure, which is defined by the combination of static topology and dynamic origin-destination~(OD) flow:
\begin{equation}
    \mathcal{L}_{\mathrm{sem}} = \frac{1}{N_S^2} \left\| \hat{\bm{H}}_S\hat{\bm{H}}_S^\top - \left( \lambda \bm{A}_S + (1-\lambda) \bm{O}_S \right) \right\|_F^2,
\end{equation}
where $\bm{O}_S$ denotes the normalized OD matrix derived from trajectory data, and $\lambda$ is a balancing coefficient.

\subsubsection{Entropy Loss.} The preservation of low-frequency components in hierarchical modeling relies on the assignment matrices satisfying theoretical constraints. To promote this, we minimize the entropy of the assignment distributions:
\begin{equation}
    \mathcal{L}_{\mathrm{ent}}^{SL} = -\frac{1}{N_S} \sum_{i=1}^{N_S} \sum_{j=1}^{N_L} \bm{A}_{SL}[i,j] \log(\bm{A}_{SL}[i,j]),
    \nonumber
\end{equation}
\begin{equation}
    \mathcal{L}_{\mathrm{ent}}^{LR} = -\frac{1}{N_L} \sum_{j=1}^{N_L} \sum_{k=1}^{N_R} \bm{A}_{LR}[j,k] \log(\bm{A}_{LR}[j,k]),
    \nonumber
\end{equation}
\begin{equation}
    \mathcal{L}_{\mathrm{ent}} = \frac{1}{2}\left(\mathcal{L}_{\mathrm{ent}}^{SL}+\mathcal{L}_{\mathrm{ent}}^{LR}\right).
\end{equation}
This encourages the assignments to be sharp, ensuring clearer hierarchical structure.

Finally, the overall loss function is formulated as
\begin{equation}
    \mathcal{L} = \gamma_1 \mathcal{L}_{\mathrm{align}} + 
    \gamma_2 \mathcal{L}_{\mathrm{rec}} + 
    \gamma_3 \mathcal{L}_{\mathrm{sem}} + 
    \gamma_4 \mathcal{L}_{\mathrm{ent}},
\end{equation}
where $\gamma_1$, $\gamma_2$, $\gamma_3$, and $\gamma_4$ are hyperparameters that control the relative contributions of each term.

\section{Experiments}
In this section, we conduct experiments to demonstrate the effectiveness of our proposed model.

\subsection{Experimental Setup}

\subsubsection{Construction of the Datasets.}
To evaluate the performance of our model, we use three real-world public datasets collected from Beijing~(\emph{BJ}), Chengdu~(\emph{CD}), and Xi'an~(\emph{XA}), which are major metropolitan areas in China.
For all datasets, we collect road network information from \emph{OpenStreetMap}~\footnote{\url{https://www.openstreetmap.org/}}.
The \emph{BJ} dataset contains taxi trajectory data sampled every minute, while the \emph{CD} and \emph{XA} datasets are sampled every 2--4 seconds. 
We perform map matching~\cite{yang2018fast} by aligning GPS points to road segments, which transforms the trajectory data into segment sequences.
We then split the sequences into individual trajectories using the provided boundary indicators. 
For all downstream tasks, we divide each dataset into training, validation, and test sets with a ratio of 7:1:2.
A detailed description is provided in Appendix~B.

\begin{table*}
    \footnotesize
    \centering
    \caption{Performance comparison across four tasks on three datasets. Higher is better for all metrics except EDT. \textbf{Bold} indicates the best result, and \underline{underline} indicates the second-best.
    }
    \begin{tabular}{c|c|c|@{\hspace{3pt}}c@{\hspace{3pt}}c@{\hspace{3pt}}c@{\hspace{3pt}}|@{\hspace{3pt}}c@{\hspace{3pt}}c@{\hspace{3pt}}c@{\hspace{3pt}}c@{\hspace{3pt}}c@{\hspace{3pt}}|@{\hspace{3pt}}c@{\hspace{3pt}}c@{\hspace{3pt}}c@{\hspace{3pt}}|c}
    \toprule
    Task & Dataset & Metric & \makecell{DeepWalk} & IRN2vec & Toast & GCN & DGI & \makecell{Geom-GCN} & \makecell{DiffPool} & HRNR & GT & \makecell{Graph-\\ormer} & \makecell{Node-\\Former} & HiFiNet \\
    \midrule
    \multirow{6}{*}{\rotatebox{90}{\makecell{Next Location\\ Prediction}}}
    & \multirow{2}{*}{BJ}
    & ACC@1$\uparrow$ & 0.383 & 0.371 & 0.391 & 0.387 & 0.381 & 0.391 & 0.398 & \underline{0.412} & 0.362 & 0.374 & 0.371 & \textbf{0.426} \\
    & & ACC@5$\uparrow$ & 0.527 & 0.498 & 0.542 & 0.517 & 0.535 & 0.526 & 0.532 & \underline{0.556} & 0.483 & 0.511 & 0.502 & \textbf{0.587} \\
    & \multirow{2}{*}{CD}
    & ACC@1$\uparrow$ & 0.403 & 0.324 & 0.369 & 0.388 & 0.390 & 0.398 & 0.409 & \underline{0.420} & 0.379 & 0.383 & 0.381 & \textbf{0.442} \\
    & & ACC@5$\uparrow$ & 0.556 & 0.454 & 0.542 & 0.552 & 0.538 & 0.546 & 0.556 & \underline{0.571} & 0.506 & 0.516 & 0.515 & \textbf{0.665} \\
    & \multirow{2}{*}{XA}
    & ACC@1$\uparrow$ & 0.346 & 0.324 & 0.335 & 0.333 & 0.375 & 0.346 & 0.358 & \underline{0.376} & 0.316 & 0.310 & 0.318 & \textbf{0.399} \\
    & & ACC@5$\uparrow$ & 0.461 & 0.457 & 0.460 & 0.455 & 0.480 & 0.476 & 0.487 & \underline{0.500} & 0.456 & 0.457 & 0.475 & \textbf{0.546} \\
    \midrule
    \multirow{6}{*}{\rotatebox{90}{\makecell{Label\\Classification}}}
    & \multirow{2}{*}{BJ}
    & F1$\uparrow$ & 0.676 & 0.733 & 0.679 & 0.790 & 0.797 & 0.773 & 0.769 & 0.819 & 0.821 & 0.817 & \underline{0.823} & \textbf{0.838} \\
    & & AUC$\uparrow$ & 0.825 & 0.836 & 0.825 & 0.849 & 0.861 & 0.846 & 0.831 & 0.885 & 0.883 & 0.882 & \underline{0.887} & \textbf{0.906} \\
    & \multirow{2}{*}{CD} 
    & F1$\uparrow$ & 0.702 & 0.686 & 0.645 & 0.716 & 0.726 & 0.719 & 0.702 & 0.747 & 0.763 & 0.752 & \underline{0.772} & \textbf{0.796} \\
    & & AUC$\uparrow$ & 0.721 & 0.706 & 0.712 & 0.734 & 0.737 & 0.735 & 0.735 & 0.782 & 0.805 & 0.798 & \underline{0.835} & \textbf{0.869} \\
    & \multirow{2}{*}{XA}
    & F1$\uparrow$ & 0.626 & 0.627 & 0.640 & 0.646 & 0.659 & 0.651 & 0.650 & 0.694 & \underline{0.705} & 0.688 & 0.701 & \textbf{0.720} \\
    & & AUC$\uparrow$ & 0.639 & 0.629 & 0.653 & 0.661 & 0.672 & 0.665 & 0.686 & 0.716 & \underline{0.736} & 0.708 & 0.733 & \textbf{0.811} \\
    \midrule
    \multirow{6}{*}{\rotatebox{90}{\makecell{Destination\\Prediction}}}
    & \multirow{2}{*}{BJ}
    & ACC@1$\uparrow$ & 0.229 & 0.215 & 0.271 & 0.232 & 0.262 & 0.242 & 0.242 & \underline{0.277} & 0.273 & 0.270 & 0.275 & \textbf{0.297} \\
    & & ACC@5$\uparrow$ & 0.321 & 0.313 & 0.396 & 0.352 & 0.366 & 0.357 & 0.365 & \underline{0.401} & 0.396 & 0.392 & 0.399 & \textbf{0.428} \\
    & \multirow{2}{*}{CD}
    & ACC@1$\uparrow$ & 0.187 & 0.235 & 0.239 & 0.251 & 0.171 & 0.267 & 0.270 & 0.281 & 0.282 & 0.282 & \underline{0.284} & \textbf{0.295} \\
    & & ACC@5$\uparrow$ & 0.362 & 0.346 & 0.342 & 0.377 & 0.321 & 0.394 & 0.393 & 0.407 & 0.405 & 0.403 & \underline{0.409} & \textbf{0.426} \\
    & \multirow{2}{*}{XA}
    & ACC@1$\uparrow$ & 0.167 & 0.210 & 0.198 & 0.217 & 0.175 & 0.226 & 0.232 & 0.256 & 0.254 & 0.248 & \underline{0.260} & \textbf{0.291} \\
    & & ACC@5$\uparrow$ & 0.289 & 0.305 & 0.306 & 0.334 & 0.315 & 0.352 & 0.353 & 0.375 & 0.364 & 0.361 & \underline{0.379} & \textbf{0.430} \\
    \midrule
    \multirow{6}{*}{\rotatebox{90}{\makecell{Route\\Planning}}}
    & \multirow{2}{*}{BJ}
    & F1$\uparrow$ & 0.304 & 0.287 & \underline{0.325} & 0.299 & 0.297 & 0.305 & 0.305 & 0.324 & 0.301 & 0.312 & 0.303 & \textbf{0.339} \\
    & & EDT$\downarrow$ & 8.151 & 8.853 & 7.899 & 8.241 & 8.114 & 8.138 & 8.142 & \underline{7.833} & 8.232 & 8.083 & 8.108 & \textbf{7.773} \\
    & \multirow{2}{*}{CD}
    & F1$\uparrow$ & 0.319 & 0.316 & \underline{0.390} & 0.335 & 0.339 & 0.330 & 0.346 & 0.374 & 0.332 & 0.345 & 0.343 & \textbf{0.498} \\
    & & EDT$\downarrow$ & 8.006 & 8.011 & \underline{7.294} & 7.869 & 7.818 & 7.737 & 7.655 & 7.350 & 7.773 & 7.673 & 7.715 & \textbf{7.171} \\
    & \multirow{2}{*}{XA}
    & F1$\uparrow$ & 0.276 & 0.259 & \underline{0.332} & 0.282 & 0.286 & 0.281 & 0.291 & 0.321 & 0.280 & 0.291 & 0.289 & \textbf{0.377} \\
    & & EDT$\downarrow$ & 8.585 & 9.157 & 8.205 & 8.866 & 8.611 & 8.621 & 8.519 & \underline{8.130} & 8.732 & 8.529 & 8.659 & \textbf{7.980} \\
    \bottomrule
    \end{tabular}
    \label{tab:result}
\end{table*}

\subsubsection{Methods to Compare.} 
In our experiments, we consider three types of baselines for a comprehensive comparison:

\textbullet \textit{Random Walk-based Models}:  
These methods learn node embeddings by generating random walk sequences on the graph and applying shallow embedding techniques. Representative baselines include \emph{DeepWalk}~\cite{perozzi2014deepwalk}, \emph{IRN2Vec}~\cite{wang2019learning}, and \emph{Toast}~\cite{chen2021robust}.

\textbullet \textit{GNN-based Models}:  
These baselines leverage message passing neural networks to aggregate local or hierarchical information from neighbors. We consider \emph{GCN}~\cite{kipf2016semi}, \emph{DGI}~\cite{velickovic2019deep}, \emph{Geom-GCN}~\cite{pei2018geom}, \emph{DiffPool}~\cite{ying2018hierarchical}, and \emph{HRNR}~\cite{wu2020learning}.

\textbullet \textit{Graph Transformer-based Models}:  
These methods apply transformer architectures to graph data to capture both local structure and global dependencies. We include \emph{GT}~\cite{dwivedi2020generalization}, \emph{Graphormer}~\cite{ying2021transformers}, and \emph{NodeFormer}~\cite{wu2022nodeformer}.

Due to space limitations, details of each baseline and their adaptations to road networks are provided in Appendix~C.

\subsubsection{Evaluation Tasks.}
We evaluate the learned road segment representations on four traffic-related tasks: (1) next location prediction, which aims to predict the next road segment based on historical trajectories; (2) label classification, where each segment is assigned a semantic label (e.g., bridge, tunnel); (3) destination prediction, which infers the final destination from a partial trajectory; and (4) route planning, which reconstructs the full path between a source and destination. Detailed setups are deferred to Appendix~D.

\subsubsection{Evaluation Metrics.}
We use task-specific evaluation metrics as follows. 
For next location and destination prediction, we treat them as ranking tasks and report top-1 and top-5 accuracy, denoted by \emph{ACC@1} and \emph{ACC@5}. 
For label classification, we report \emph{F1-score} and \emph{AUC}. The former balances precision and recall of binary classification, and the latter computes the area under the  ROC curve.
For route planning, we evaluate the predicted route $r'$ and the actual route $r$, which share the same source and destination. 
We compute \emph{F1-score} based on overlapping locations: $P = \frac{|r \cap r'|}{|r'|}$, 
$R = \frac{|r \cap r'|}{|r|}$, and 
$F1 = \frac{2PR}{P + R}$. Additionally, we report the edit distance~(\emph{EDT}), which measures the minimum number of edit operations required to transform $r'$ into $r$.

\subsection{Results and Analysis}

Table~\ref{tab:result} summarizes the performance of all baselines and our proposed model across four tasks and three datasets.

First, random walk-based methods perform poorly on label classification and destination prediction, but show relatively better results on next location prediction and route planning. This is likely due to their reliance on local graph topology, which aligns well with tasks that emphasize short-range transitions. Toast, an enhanced variant incorporating trajectory information, achieves strong performance on route planning. However, these methods fail to capture node-level attributes and long-range dependencies, limiting their generalization on more semantic or global tasks.

Second, GNN-based models generally outperform random walk-based ones, benefiting from their ability to integrate graph features and learn node interactions through deep message passing. Nevertheless, models such as GCN and DGI still struggle with long-range reasoning. Geom-GCN improves representation learning by introducing spatial priors, while DiffPool extends the receptive field via hierarchical pooling. Among these, HRNR achieves the strongest performance by leveraging dual semantic-guided assignment matrices to enhance hierarchical representation.

Third, graph transformer-based models demonstrate the opposite performance pattern to random walk-based methods: they perform better on label classification and destination prediction, but worse on next location prediction and route planning. This can be attributed to their global attention mechanisms, which favor tasks requiring holistic semantic context, but may over-smooth or overlook local connectivity patterns critical for sequential prediction.

Finally, our proposed model HiFiNet consistently outperforms all baselines across tasks and datasets. By incorporating a hierarchical architecture and frequency-aware decomposition, HiFiNet effectively captures both local and global structures. The explicit separation of low- and high-frequency components preserves multi-scale semantics, enabling more expressive and generalizable representations for various downstream tasks.

\begin{figure}
    \centering
    \includegraphics[width=1\linewidth]{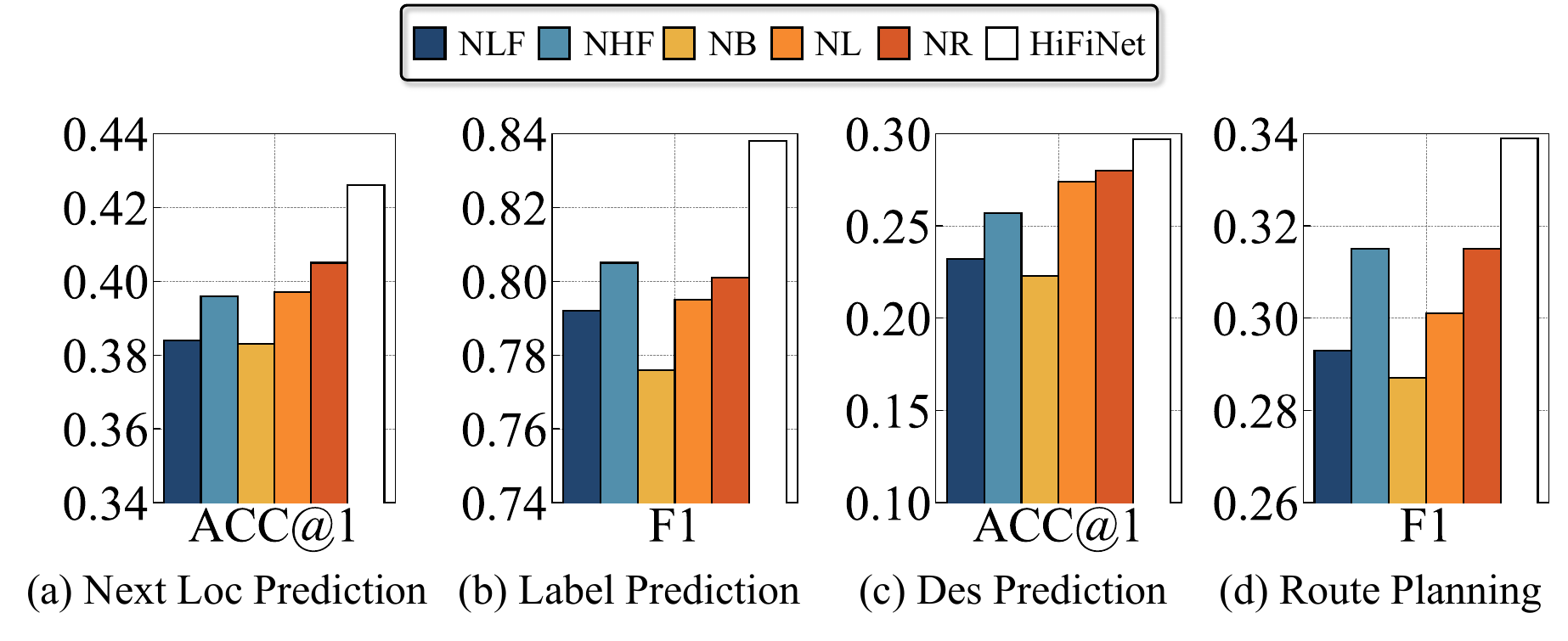}
    \caption{Ablation study of our model on Beijing dataset.}
    \label{fig:ablation}
\end{figure}

\subsection{Ablation Study}

HiFiNet contains two key components: a hierarchical architecture and a frequency-decomposition module. We design five model variants to assess the contribution of each component, and conduct ablation studies on the \emph{Beijing} dataset (similar trends are observed on others and omitted for brevity): (1) \underline{\emph{NL}}: without the locality level; (2) \underline{\emph{NR}}: without the region level; (3) \underline{\emph{NB}}: without both locality and region levels; (4) \underline{\emph{NLF}}: without the low-frequency component; and (5) \underline{\emph{NHF}}: without the high-frequency component.

As shown in Fig.~\ref{fig:ablation}, the performance ranking is: \underline{\emph{NB}} $<$ \underline{\emph{NL}} $<$ \underline{\emph{NR}} $<$ \emph{HiFiNet} and \underline{\emph{NLF}} $<$ \underline{\emph{NHF}} $<$ \emph{HiFiNet}. 
The first group validates the role of hierarchy—removing both levels leads to the largest drop, and localities contribute more than regions, likely due to their finer granularity.
The second group highlights the effectiveness of frequency decomposition. Removing either frequency component degrades performance, confirming their complementarity. Low-frequency components appear more critical, as they encode global structural semantics essential for context-aware tasks, while high-frequency features enhance 
fine-grained local discrimination.

\begin{figure}[!t]
  \centering
  \subfloat[The number of localities]{
    \includegraphics[width=0.23\textwidth]{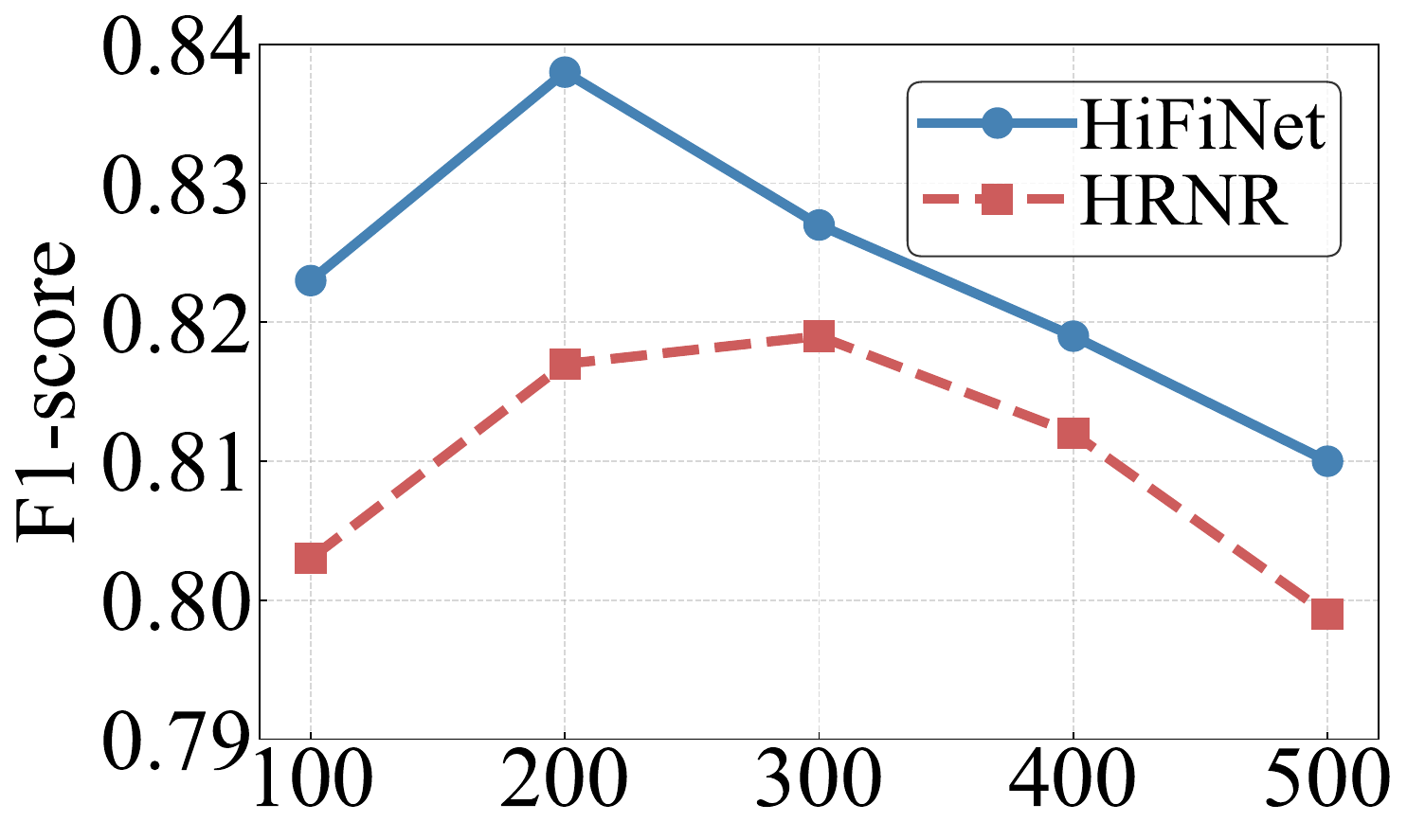}%
    \label{locality_num}}
  \subfloat[The number of regions]{
    \includegraphics[width=0.23\textwidth]{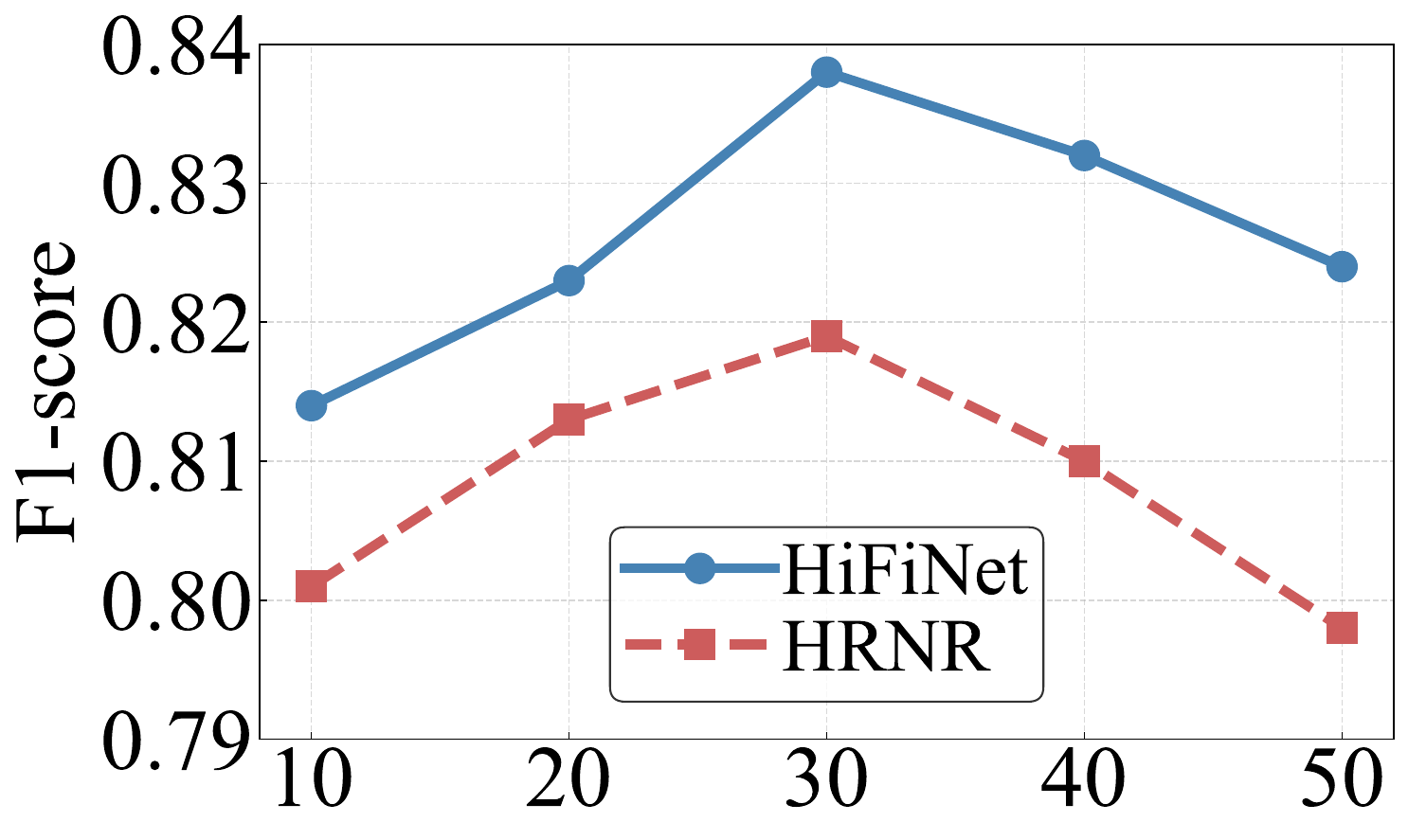}%
    \label{region_num}}
  \caption{Parameter sensitivity of our model on Beijing dataset on label classification task.}
  \label{fig:parameter}
\end{figure}

\subsection{Parameter Sensitivity}
In addition to model components, several hyperparameters require tuning in our model. We conduct sensitivity analyses on the \emph{Beijing} dataset on the label classification task. The complete experimental results are provided in Appendix~E.

Specifically, we vary the number of locality nodes $N_L$ in $\{100, 200, 300, 400, 500\}$ and the number of region nodes $N_R$ in $\{10, 20, 30, 40, 50\}$. As shown in Fig.~\ref{fig:parameter}, the model performs best when $N_L = 200$ and $N_R = 30$. We observe that performance initially improves as the number of locality or region nodes increases, due to finer-grained representations and improved structural abstraction. However, overly large values lead to degraded performance, possibly due to increased noise or over-segmentation. Since regions are formed by aggregating finer-grained localities, it is reasonable to use more localities than regions. Overall, the model shows stable performance across a wide range of parameter settings, highlighting its robustness and applicability.

\subsection{Qualitative Analysis}

\begin{figure}
    \centering
    \includegraphics[width=1\linewidth]{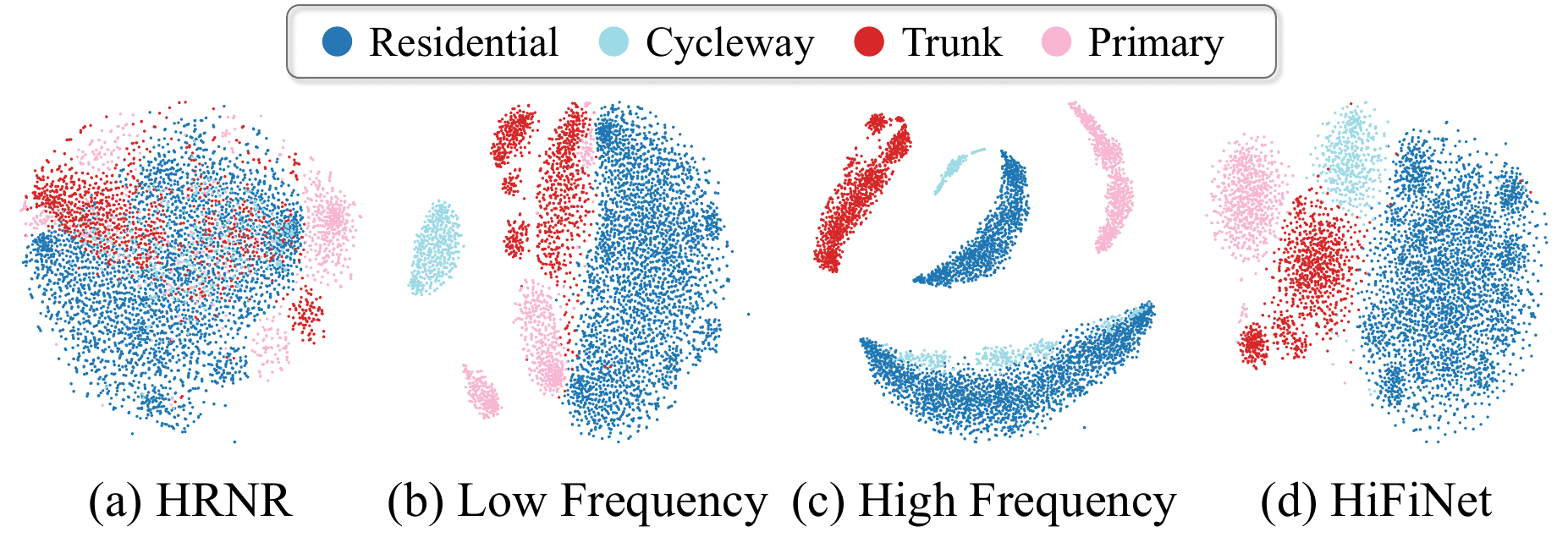}
    \caption{The t-SNE visualization of road network representations under different frequency configurations.}
    \label{fig:qualitative_analysis}
\end{figure}

To evaluate the representational quality of HiFiNet, we visualize the t-SNE~\cite{maaten2008visualizing} projections of road segment embeddings under different frequency configurations, as shown in Fig.~\ref{fig:qualitative_analysis}. The baseline model HRNR produces entangled embeddings with no clear separation between road types.
In contrast, HiFiNet's low-, high-, and fused-frequency representations exhibit clearer structural distinctions. For road types such as \textit{residential} and \textit{cycleway}, which have smoother and more regular patterns, low-frequency features form compact clusters. Their high-frequency counterparts, however, tend to fragment due to local variability and noise.
Conversely, for more dynamic road types like \textit{trunk} and \textit{primary}, high-frequency features better capture structural complexity, while low-frequency features result in overlap or dispersion.
Overall, the fused representation combines global and local information, yielding distinct and semantically coherent clusters across all road types. These results qualitatively demonstrate the advantage of frequency-aware hierarchical modeling in capturing multi-scale road network structures.

\section{Conclusion}

In this paper, we introduce HiFiNet, a novel framework that unifies spatial and spectral modeling for road network representation learning. By constructing a multi-level hierarchical graph, HiFiNet enables localized frequency decomposition, capturing both coarse spatial semantics and fine-grained spectral variations. We propose a decomposition–updating–reconstruction paradigm to explicitly model low- and high-frequency components and integrate them into expressive node representations.
We theoretically demonstrate that the hierarchical projection naturally acts as a spectral low-pass filter, separating frequency components and mitigating over-smoothing. Experiments on multiple real-world datasets and tasks demonstrate the robustness and generalization of HiFiNet.
Beyond road networks, our unified spatial-spectral approach offers new insights for graph learning in broader spatio-temporal domains. We hope this work inspires future research into frequency-aware models that leverage structured hierarchies to better align spatial and spectral perspectives.

\section*{Acknowledgments}
Jingyuan Wang's work was partially supported by the National Natural Science Foundation of China (No. 72222022, 72171013, 72242101) and the Fundamental Research Funds for the Central Universities (JKF-2025017226182). Leong Hou U's work was partially supported by the Science and Technology Development Fund Macau SAR (0003/2023/RIC, 0052/2023/RIA1, 0031/2022/A, 001/2024/SKL for SKL-IOTSC).

\bibliography{aaai2026}

\appendix

\section{Proofs of Theoretical Results}
\label{app:proofs}

This appendix provides the detailed theoretical support for the spectral properties of the proposed hierarchical structure. We first present two supporting lemmas and then prove the main theorem.

\begin{lemma}
\label{lemma:laplacian_projection}
Let \( A_{XY} \in \mathbb{R}^{N_Y \times N_X} \) be a hard assignment matrix that satisfies equi-partition and row-normalization conditions. Let \( L_X = D_X - A_X \) denote the Laplacian of the original graph, and let \( A_Y = A_{XY} A_X A_{XY}^\top \) be the adjacency of the coarsened graph. Then the Laplacian of the coarsened graph satisfies:
\[
L_Y = A_{XY} L_X A_{XY}^\top.
\]
\end{lemma}

\begin{proof}
By definition, the Laplacian of the coarsened graph is
\[
L_Y = D_Y - A_Y = D_Y - A_{XY} A_X A_{XY}^\top.
\]
Besides, we have
\begin{align*}
A_{XY} L_X A_{XY}^\top
&= A_{XY} (D_X-A_X) A_{XY}^\top \\
&= A_{XY} D_X A_{XY}^\top-A_{XY} A_X A_{XY}^\top.
\end{align*}
Thus, to show \( L_Y = A_{XY} L_X A_{XY}^\top \), it suffices to prove:
\[
D_Y = A_{XY} D_X A_{XY}^\top.
\]

We now consider the diagonal entry \( [D_Y]_{ii} \), which is the degree of the \( i \)-th node in coarsened graph \( Y \):
\begin{align*}
[D_Y]_{ii} 
&= \sum_j [A_Y]_{ij} = \sum_j [A_{XY} A_X A_{XY}^\top]_{ij} \\
&= \sum_j \sum_k \sum_l [A_{XY}]_{ik} [A_X]_{kl} [A_{XY}^T]_{lj} \\
&= \sum_j \sum_k \sum_l [A_{XY}]_{ik} [A_X]_{kl} [A_{XY}]_{jl} \\
&= \sum_k [A_{XY}]_{ik} \left( \sum_l [A_X]_{kl} \sum_j [A_{XY}]_{jl} \right).
\end{align*}

Assume each coarse node \( \mathcal{C}_i \) contains \( m \) original nodes. Since \( A_{XY} \) is an assignment matrix with equi-partition and row-normalization, we have
\[
\sum_j [A_{XY}]_{jl} = \frac{1}{\sqrt{m}} \quad \text{if } l \in \mathcal{C}_i.
\]
And since the degree of node \( k \) in the original graph is
\[
\sum_l [A_X]_{kl} = [D_X]_{kk},
\]
we get
\[
[D_Y]_{ii} = \frac{1}{\sqrt{m}} \sum_k [A_{XY}]_{ik} [D_X]_{kk} = \frac{1}{m} \sum_{k \in \mathcal{C}_i} [D_X]_{kk}.
\]

Now we compute the corresponding diagonal entry in \( A_{XY} D_X A_{XY}^\top \):
\begin{align*}
[A_{XY} D_X A_{XY}^\top]_{ii} 
&= \sum_k [A_{XY}]_{ik} [D_X]_{kk} [A_{XY}]_{ik} \\
&= \sum_k ([A_{XY}]_{ik})^2 [D_X]_{kk}.
\end{align*}
Since \( A_{XY} \) is hard assignment with row-normalization,
\[
[A_{XY}]_{ik} = \frac{1}{\sqrt{m}} \text{ if } k \in \mathcal{C}_i,\quad 0 \text{ otherwise}.
\]
So we obtain
\[
[A_{XY} D_X A_{XY}^\top]_{ii} = \frac{1}{m} \sum_{k \in \mathcal{C}_i} [D_X]_{kk} = [D_Y]_{ii}.
\]

Hence, we have
\[
D_Y = A_{XY} D_X A_{XY}^\top,
\]
and thus
\[
L_Y = A_{XY} L_X A_{XY}^\top.
\]
\end{proof}

\begin{lemma}
\label{lemma:energy_contraction}
Let \( A_{XY} \in \mathbb{R}^{N_Y \times N_X} \) be a hard assignment matrix satisfying equi-partition and row-normalization. Let \( \mathbf{z}_X \in \mathbb{R}^{N_X} \) be any graph signal on the original graph with Laplacian \( L_X \), and define its projection \( \mathbf{z}_Y = A_{XY} \mathbf{z}_X \in \mathbb{R}^{N_Y} \). Then the Dirichlet energy satisfies:
\[
\mathbf{z}_Y^\top L_Y \mathbf{z}_Y \leq \mathbf{z}_X^\top L_X \mathbf{z}_X.
\]
\end{lemma}

\begin{proof}
From Lemma~\ref{lemma:laplacian_projection}, we have
\[
L_Y = A_{XY} L_X A_{XY}^\top.
\]
Thus,
\[
\begin{aligned}
\mathbf{z}_Y^\top L_Y \mathbf{z}_Y &= \mathbf{z}_X^\top A_{XY}^\top (A_{XY} L_X A_{XY}^\top) A_{XY} \mathbf{z}_X \\
&= \mathbf{z}_X^\top (A_{XY}^\top A_{XY}) L_X (A_{XY}^\top A_{XY}) \mathbf{z}_X.
\end{aligned}
\]
Since \( A_{XY}^\top A_{XY} \preceq I \) under the equi-partition and row-normalization assumptions,\footnote{This can be shown by observing that each row of \( A_{XY} \) is a hard cluster assignment that spreads evenly over \( m \) nodes, so the resulting product is a contraction.} we obtain
\[
(A_{XY}^\top A_{XY}) L_X (A_{XY}^\top A_{XY}) \preceq L_X
\]
Hence,
\[
\mathbf{z}_Y^\top L_Y \mathbf{z}_Y \leq \mathbf{z}_X^\top L_X \mathbf{z}_X.
\]
\end{proof}

\begin{theorem}
\label{thm:spectral_filtering}
Let \( A_{XY} \in \mathbb{R}^{N_Y \times N_X} \) denote an assignment matrix satisfying the equi-partition and row-normalization properties. Then, the projection of graph signals from the original graph \( X \) to the coarsened graph \( Y \) approximately preserves the low-frequency energy while attenuating high-frequency components.
\end{theorem}

\begin{proof}
Let the total Dirichlet energy of a signal \( \mathbf{z}_X \) on the original graph be decomposed into low- and high-frequency parts:
\[
E_X = E_X^{\mathrm{low}} + E_X^{\mathrm{high}}.
\]
% Then, under the projection \( \mathbf{z}_Y = A_{XY} \mathbf{z}_X \), the coarsened signal satisfies:
% \[
% E_Y^{\mathrm{low}} \approx E_X^{\mathrm{low}}, \quad E_Y^{\mathrm{high}} < E_X^{\mathrm{high}}.
% \]
From Lemma~\ref{lemma:energy_contraction}, we have
\[
E_Y = \mathbf{z}_Y^\top L_Y \mathbf{z}_Y \leq \mathbf{z}_X^\top L_X \mathbf{z}_X = E_X.
\]

Let \( \{\mathbf{u}_j\}_{j=1}^{N_X} \) be an orthonormal set of eigenvectors of \( L_X \) with corresponding eigenvalues \( \lambda_j \), such that
\[
L_X \mathbf{u}_j = \lambda_j \mathbf{u}_j, \quad \text{and} \quad \mathbf{z}_X = \sum_{j=1}^{N_X} \alpha_j \mathbf{u}_j.
\]
Then the Dirichlet energy of \( \mathbf{z}_X \) is
\[
E_X = \mathbf{z}_X^\top L_X \mathbf{z}_X = \sum_{j=1}^{N_X} \alpha_j^2 \lambda_j.
\]

Now consider a low-frequency eigenvector \( \mathbf{u}_j \in \mathbb{R}^{N_X} \) of the original Laplacian \( L_X \), with eigenvalue \( \lambda_j \ll 1 \), and assume \( \mathbf{u}_j \) is smooth over clusters—that is, nearly constant within each cluster. Then the projection via \( A_{XY} \) gives
\[
\mathbf{v}_j := A_{XY} \mathbf{u}_j \in \mathbb{R}^{N_Y}, \quad \text{with} \quad \| \mathbf{v}_j \|^2 \approx \| \mathbf{u}_j \|^2 = 1,
\]
since \( A_{XY} \) is row-normalized and distributes node weights evenly (equi-partition). Moreover, under Lemma~\ref{lemma:laplacian_projection}, the Laplacian of the coarsened graph satisfies
\[
L_Y = A_{XY} L_X A_{XY}^\top.
\]
Then the Dirichlet energy of the projected low-frequency component is
\begin{align*}    
E_{Y,j}^{\text{low}} 
&:= \mathbf{v}_j^\top L_Y \mathbf{v}_j \\
&= \mathbf{u}_j^\top A_{XY}^\top L_Y A_{XY} \mathbf{u}_j \\
&= \mathbf{u}_j^\top A_{XY}^\top A_{XY} L_X A_{XY}^\top A_{XY} \mathbf{u}_j.
\end{align*}
Let \( P := A_{XY}^\top A_{XY} \preceq I \) be the contraction matrix. Then
\[
E_{Y,j}^{\text{low}} = \mathbf{u}_j^\top P L_X P \mathbf{u}_j \approx \lambda_j,
\]
because \( P \mathbf{u}_j \approx \mathbf{u}_j \) for smooth \( \mathbf{u}_j \in \mathrm{im}(P) \).

On the original graph, the energy of this component is
\[
E_{X,j}^{\text{low}} := \mathbf{u}_j^\top L_X \mathbf{u}_j = \lambda_j.
\]
Therefore, we have
\[
E_{Y,j}^{\text{low}} \approx E_{X,j}^{\text{low}}.
\]

Let \( \mathcal{F}_{\text{low}} := \{ j \mid \lambda_j \leq \epsilon \} \) for some small threshold \( \epsilon > 0 \) denote low-frequency components.
Summing over all such low-frequency components \( \mathcal{F}_{\text{low}} \), we have
\[
E_Y^{\text{low}} = \sum_{j \in \mathcal{F}_{\text{low}}} E_{Y,j}^{\text{low}} \approx \sum_{j \in \mathcal{F}_{\text{low}}} E_{X,j}^{\text{low}} = E_X^{\text{low}}.
\]

Finally, combining this with the total energy inequality \( E_Y \leq E_X \), and \( E_Y^{\text{low}} \approx E_X^{\text{low}} \), we conclude that
\[
E_Y^{\text{high}} = E_Y - E_Y^{\text{low}} < E_X - E_X^{\text{low}} = E_X^{\text{high}}.
\]
That is, high-frequency components are attenuated by the projection.
\end{proof}

\section{Dataset Description}
\label{app:dataset}

\begin{table}[t]
    \footnotesize
    \centering
    \caption{Statistics of the three datasets after preprocessing.}
    \begin{tabular}{c|c|c|c}
        \toprule
        Statistics &  Beijing & Chengdu & Xi'an \\
        \midrule
        \#~road segments & 15,042 & 2,857 & 3,686 \\
        \#~road edges & 47,082 & 8,224 & 7,341 \\
        \#~road types & 17 & 13 & 12 \\
        % \#~road labels & 708 & 303 & 291 \\
        graph diameter& 131 & 71 & 47 \\
        \midrule
        \#~driving records & 16,040,662 &  9,632,481 & 6,672,027 \\
        \#~trajectories & 302,654 & 224,184 & 493,254 \\
        average hop number& 48 & 35 & 28 \\
        \bottomrule
    \end{tabular}
    \label{tab:statistic}
\end{table}

We utilize three real-world road network and trajectory datasets in our experiments.  
Table~\ref{tab:statistic} summarizes the statistics of the datasets after preprocessing.  
All three road networks exhibit long graph diameters, and notably, the average hop distance between road segments is also considerably large, reflecting the complex topology of urban-scale transportation systems.

Specifically, we extract road network information from \emph{OpenStreetMap}~\footnote{https://www.openstreetmap.org/} for all three datasets.  
In our setting, only road segments are considered, while other geographic locations (e.g., intersections or off-network points) are excluded.  
After obtaining the road network data, we perform map matching using the open-source tool \emph{FMM}~\footnote{\url{https://www.github.com/cyang-kth/fmm}}.  
The purpose of map matching is to align sampled GPS points with corresponding road segments in the network, thereby transforming raw GPS trajectories into time-ordered sequences of road segments.  
To reduce redundancy due to high-frequency sampling, we remove consecutive points that are mapped to the same road segment and retain only the entry and exit points for each segment.

The \emph{Beijing Taxi} dataset was collected from over 18{,}000 taxis operating in Beijing, China, during the period from November 1 to November 30, 2011.  
Each trajectory record is represented as a tuple $\langle \mathit{tid}, \mathit{te}, \mathit{longitude}, \mathit{latitude}, \mathit{state} \rangle$,  
where $\mathit{tid}$ denotes the unique identifier of a taxi, $\mathit{te}$ is the timestamp, and $\mathit{state}$ indicates whether the taxi is carrying a passenger at time $\mathit{te}$.  
The state information allows us to segment continuous records into individual trips, where the state ``\emph{No passengers}'' marks the end of a trip.
The \emph{Chengdu Taxi} and \emph{Xi‘an Taxi} datasets are public trajectory datasets released by the \emph{DiDi GAIA Open Dataset} platform.  
Each dataset contains one month of complete trajectory data (from November 1 to November 30, 2016) for all DiDi-operated taxis running within the second ring roads of Xian and Chengdu, respectively.  
Each individual ride is treated as a separate trajectory in the dataset.

% \begin{figure}[t]
%   \centering
%   \subfloat[Next Location Prediction]{
%     \includegraphics[width=0.23\textwidth]{figures/parameter_next_location_prediction.pdf}%
%     \label{next loc pred}}
%   \subfloat[Label Prediction]{
%     \includegraphics[width=0.23\textwidth]{figures/parameter_label_prediction.pdf}%
%     \label{label pred}} \\
%   \subfloat[Destination Prediction]{
%     \includegraphics[width=0.23\textwidth]{figures/parameter_destination_prediction.pdf}
%     \label{des pred}}
%   \subfloat[Route Planning]{
%     \includegraphics[width=0.23\textwidth]{figures/parameter_route_plan.pdf}
%     \label{route plan}}
%   \caption{Parameter sensitivity of our model on Beijing dataset for four tasks.}
%   \label{fig:app_parameter}
% \end{figure}

\begin{figure}[t]
    \centering
    \includegraphics[width=1\linewidth]{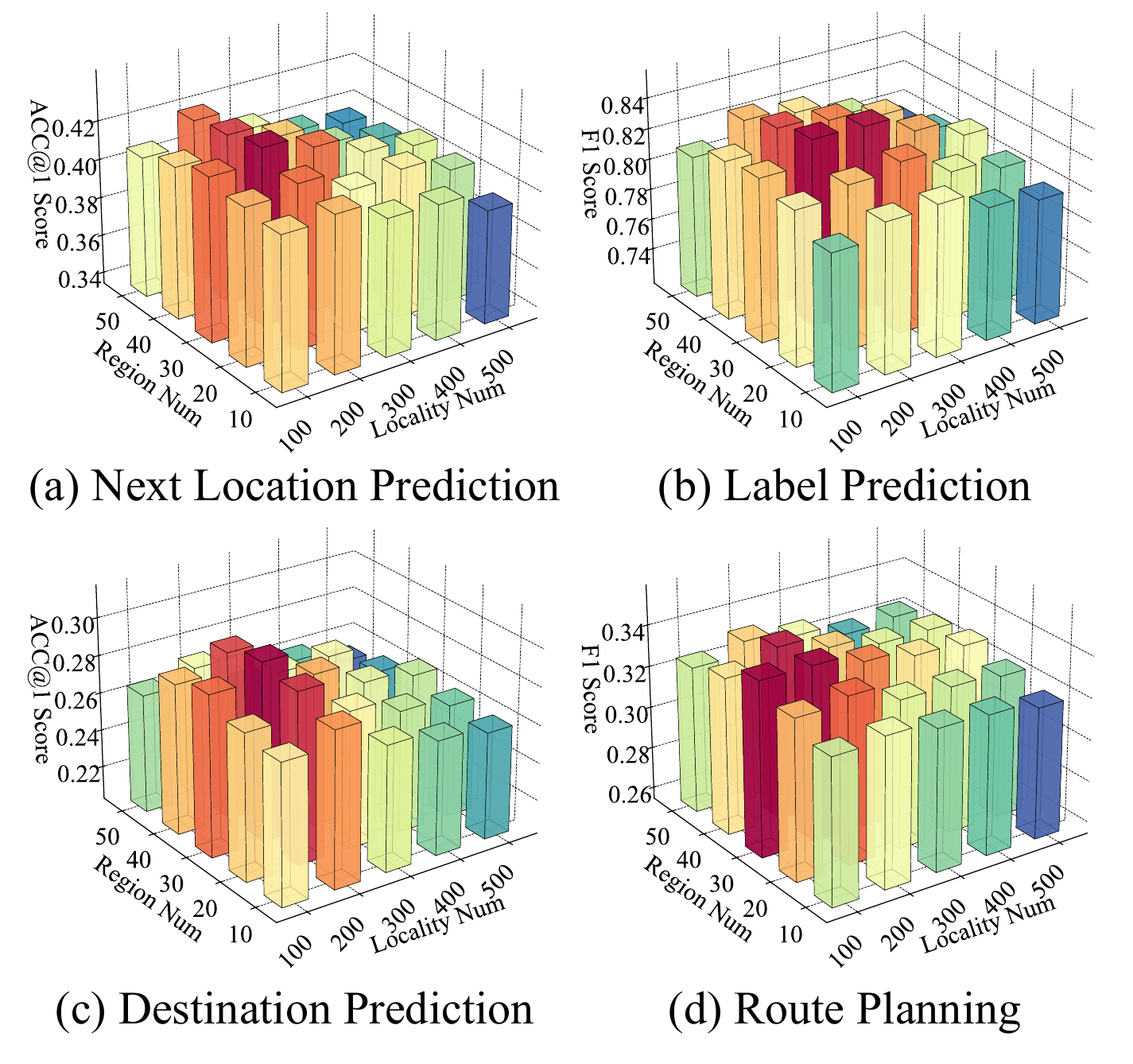}
    \caption{Parameter sensitivity of our model on Beijing dataset for four tasks.}
    \label{fig:app_parameter}
\end{figure}

\section{Baseline Model Details}
\label{app:baselines}

We provide detailed descriptions of each baseline model and how we adapt it to the road network setting:
\begin{itemize}
    \item \textbf{Random Walk-based Models}: These methods learn node embeddings by generating random walk sequences on the graph and applying shallow embedding techniques.

    \textit{DeepWalk}~\cite{perozzi2014deepwalk}:  
    DeepWalk learns latent node representations by treating truncated random walks on graphs as the equivalent of sentences in natural language models.  
    We adapt it to road networks by applying random walks on segment graphs.
    
    \textit{IRN2Vec}~\cite{wang2019learning}: 
    IRN2Vec is originally an intersection representation learning model that captures geo-locality and the mobility patterns of road users. 
    We adapt it to our setting by replacing intersections with road segments and focusing on geo-location and road type attributes.
    
    \textit{Toast}~\cite{chen2021robust}: 
    Toast is a robust embedding model for road networks that integrates an auxiliary traffic context prediction task with random walk sequences to enhance the quality of segment representations.

    \item \textbf{GNN-based Models}: These baselines leverage graph neural networks to aggregate local or hierarchical information from neighbors.

    \textit{GCN}~\cite{kipf2016semi}:  
    GCN is a foundational graph neural network model that performs layer-wise propagation based on spectral graph convolutions.  
    We apply it to road networks by learning segment-level embeddings in a supervised setting for downstream tasks.
    
    % \textit{GAT}~\cite{velickovic2017graph}:  
    % GAT is a standard graph attention network model that combines attention mechanisms with graph structure.  
    % We apply it to road networks by learning segment-level embeddings in a supervised setting for downstream tasks.
    
    \textit{DGI}~\cite{velickovic2019deep}:  
    DGI is an unsupervised graph representation learning model that maximizes the mutual information between local and global node representations to capture meaningful structural patterns.  
    We adapt DGI to road networks by using the segment graph as input and learning segment-level embeddings in an unsupervised manner for downstream tasks.
    
    \textit{Geom-GCN}~\cite{pei2018geom}:  
    Geom-GCN extends GCN by introducing a bi-level geometrical aggregation scheme to address the loss of structural information and the challenge of capturing long-range dependencies.  
    We adapt it to road networks by aggregating spatially and semantically close segments as neighbors.
    
    \textit{DiffPool}~\cite{ying2018hierarchical}: 
    DiffPool is a differentiable graph pooling model that generates hierarchical representations through learnable pooling layers.  
    We apply it with three pooling levels, without additional semantic constraints or frequency decomposition.
    
    \textit{HRNR}~\cite{wu2020learning}:  
    HRNR is a hierarchical road network representation learning framework with three levels.  
    It extends DiffPool by introducing two semantically guided assignment matrices and incorporating trajectory data to enhance representation learning.

    \item \textbf{Graph Transformer-based Models}: These methods apply transformer architectures to graph data to capture both local structure and global dependencies.
    
    \textit{GT}~\cite{dwivedi2020generalization}:  
    Graph Transformer is a generalization of the transformer architecture for arbitrary graphs, where positional encodings are given by the Laplacian eigenvectors.  
    We apply it to road networks without using edge attributes.
    
    \textit{Graphormer}~\cite{ying2021transformers}:  
    Graphormer is a transformer-based model for graphs that incorporates centrality, spatial, and edge encodings to capture both local and global dependencies.  
    We adapt it to road networks by representing road segments as nodes.
    
    \textit{NodeFormer}~\cite{wu2022nodeformer}: 
    NodeFormer introduces efficient attention mechanisms for scalable graph transformer learning by approximating softmax attention with kernel functions.  
    We apply it directly to segment graphs in road networks.
    
\end{itemize}

\section{Evaluation Tasks.}
\label{app:tasks}
We evaluate the comparison methods on four traffic-related application tasks.
For each task, we construct a simple, standard neural network architecture~(\eg~GRU or MLP) as the basic framework and incorporate the learned road network representations (mainly for road segments) as embeddings to enhance it. 
We intentionally avoid complex architectures or auxiliary data, focusing on learning generally useful road network representations while minimizing the influence of other factors.
The four application tasks are as follows:

\textbullet \textit{Next Location Prediction:} 
This task aims to predict the next location a user will visit~\cite{Wu2017Modeling}. 
We construct a GRU-based model that takes historical trajectories as input and outputs a ranked list of candidate road segments. 
To emphasize long-range dependencies, we down-sample trajectories at ten-minute intervals. 
A good method should rank the actual next location as high as possible in the candidate list.

\textbullet \textit{Label Classification:} 
This is a standard task for evaluating representation learning models~\cite{wang2019learning}. 
Our dataset provides labels for road segments (\eg~\emph{bridge}, \emph{tunnel}). 
We use a logistic regression classifier that takes segment representations as input and outputs label distributions.
The label with the highest probability is the final prediction.

\textbullet \textit{Destination Prediction:} 
This task aims to predict the final destination given a partial trajectory~\cite{xue2013destination}, which is important for tasks such as POI recommendation. 
We construct a GRU-based predictor that takes segment representations as input and predicts the destination, which is defined as the last location in the trajectory.

\textbullet \textit{Route Planning:} 
This task aims to generate the complete route connecting a source and destination location~\cite{Wei2012Constructing}, which is more challenging than next location or destination prediction. 
We design a hierarchical predictor that first encodes the observed trajectory and destination using a GRU and then predicts subsequent locations progressively. 
We treat the first and last locations in each trajectory sequence as source and destination, respectively, while the intermediate locations are hidden for prediction.

\section{Parameter Analysis}
\label{app:parameter}

In addition to model components, several hyperparameters require tuning in our model. We conduct sensitivity analyses on the \emph{Beijing} dataset across four tasks.

Specifically, we vary the number of locality nodes $N_L$ in $\{100, 200, 300, 400, 500\}$ and the number of region nodes $N_R$ in $\{10, 20, 30, 40, 50\}$. As shown in Fig.~\ref{fig:app_parameter}, the model achieves the best performance when $N_L = 200$ and $N_R = 30$. We observe that performance initially improves as the number of locality or region nodes increases, due to finer-grained representations and improved structural abstraction. However, overly large values lead to degraded performance, possibly due to increased noise or over-segmentation of the graph structure. Since regions are formed by aggregating finer-grained localities, it is reasonable for the model to use more locality nodes than region nodes. Overall, the model shows stable performance across a wide range of parameter settings, highlighting its robustness and applicability.

\end{document}